\documentclass{article}
 \usepackage{dsfont}
 \usepackage{wrapfig}

\usepackage[nonatbib,preprint]{nips_2018}
\usepackage[utf8]{inputenc} 
\usepackage[T1]{fontenc}    
\usepackage[hidelinks]{hyperref}       
\usepackage{url}            
\usepackage{booktabs}       
\usepackage{amsfonts}       
\usepackage{nicefrac}       
\usepackage{microtype}      
\usepackage{amsmath}
\usepackage{amsfonts}
\usepackage{amssymb}
\usepackage{bbold}
\usepackage{graphicx}
\usepackage{xcolor}

\title{Deep Pepper: Expert Iteration based Chess agent in the Reinforcement Learning Setting}

\author{	
  Vijaya Sai Krishna Gottipati \\
   Mila, Université de Montreal \\
 \texttt{saikrishnagv1996@gmail.com} \\
  \And 
     Kyle Goyette\thanks{Equal contribution, random order} \\
  Mila, Université de Montreal \\
 \texttt{kdgoyette@gmail.ca} \\
\And  
    Ahmad Chamseddine\footnotemark[1] \\
  École Polytechnique de Montreal \\
 \texttt{ahmad.chamseddine@polymtl.ca} \\
\And 
    Breandan Considine \\
  Mila, Université de Montreal \\
 \texttt{breandan.considine@gmail.com} \\ 
}

\begin{document}

\maketitle
\begin{abstract}
%
%
  An almost-perfect chess playing agent has been a long standing challenge in the field of Artificial Intelligence. Some of the recent advances demonstrate we are approaching that goal. In this project, we provide methods for faster training of self-play style algorithms, mathematical details of the algorithm used, various potential future directions, and discuss most of the relevant work in the area of computer chess. Deep Pepper uses embedded knowledge to accelerate the training of the chess engine over a "tabula rasa" system such as Alpha Zero. We also release our code to promote further research.
  
%
%
%
%
%

\end{abstract}

\section{Introduction and Motivation}
The game of chess has been a long standing challenge in modern artificial intelligence. While Deep Blue (\cite{deepblue}, \cite{deepblueHsu}) made headlines in 1996, the first chess engines started beating top human players in the 1970's\cite{1970chess} based on the principles laid by Von Neumann \cite{vonNeumann}, Claude Shannon\cite{Shannon} and Alan Turing\cite{Turing}.  Aided by rapid growth in computational power, computer chess has come a long way since then, with most contemporary engines using Alpha-Beta search \cite{alpha-beta} (like the Stockfish\cite{stockfish} engine, released in 2008). More recently, there has been significant progress in machine learning applied to chess, lead by DeepMind's Alpha Zero \cite{ChessZero}. Perhaps the most impressive is Alpha Zero's ability to learn the game entirely from self-play. Alpha Zero makes use of several ideas in deep learning and reinforcement learning, including function approximation with deep neural networks and more conventional heuristic search algorithms such as Monte Carlo methods. However there has been certain criticism\cite{WinNT}
claiming that Alpha Zero was given an unfair advantage against computer chess engines during evaluation, which raised questions on how we could leverage the strengths of both the approaches. We have adapted methods from Alpha Zero, deep networks and Monte Carlo Tree Search methods, with a feature representation that encodes information about the rules of chess, and the mobility of chess pieces.
%
%
%
%

Deep Pepper attempts to build on some of the principles introduced by Expert Iteration \cite{Barbernp}, Alpha Go and Alpha Zero by using deep neural network architectures, paired with Monte Carlo Tree Search, but to also diverge from Alpha Zero by avoiding a full board representation, using a hand-crafted feature vector, with the existing Stockfish engine and few other methods to accelerate training. We also provide mathematical details of the algorithm as an extension of Classification based Approximate Policy Iteration (CAPI)\cite{FarahmandPBG14} framework, and experimental details as the training progresses.

Our code may be found at this URL: \url{https://github.com/saikrishna-1996/deep_pepper_chess}
\section{Related work}
\label{gen_inst}
Chess has been explored for more than 40 years in the world of computer science, as such there exists a large amount of related work, here we shall focus on more contemporary and machine learning(ML) driven approaches.

Some of the initial ML-based chess engines attempted to play like grandmasters by imitation learning on grandmaster games. But, they failed due to the absence of negative dataset in the case of sacrificing major pieces.
Some engines attempted to overcome this problem by hard coding in some constraints. But, they still did not achieve strong play because the target policy vector was always one-hot encoded. 
So just trying to learn a policy from given states is not ideal. However, one has to note that this might work in other games like Go, Othello, Hex or connect4. Unlike all those games, chess has a "tactical" element to it and hence the agent must look ahead server moves to decide the move at any given position. Thus started the era of planning agents in computer chess. 
%

Most of the engines use alpha-beta pruning search with value function evaluation at the leaf nodes of the tree. However, it poses certain problems. Primarily, it is difficult to determine the depth of the search. And if the value function is slightly misaligned, important nodes are overlooked.



Barber et al. \cite{Barbernp} is one of the earliest known attempts of using a completely self-play algorithm to learn a board game, demonstrating promising results on the game of Hex. David Silver et al. extended their results to the game of Go \cite{alphagozero} (whose performance exceeded their previous version \cite{AlphaGo}), and further to the game of chess \cite{ChessZero}. \cite{ELFOpenGo2018} is an attempt to replicate the results of \cite{alphagozero}. Leela Chess attempted to replicate the results of AlphaZero, but is still far from the results reported in \cite{ChessZero}. Giraffe \cite{Giraffe} is one of the first reinforcement learning based agents which performed at International Master level. Our custom feature representation is inspired by Giraffe's representation. Our pretraining is based on the evaluations provided by Stockfish \cite{stockfish} which is currently the best performing open source engine.

Recent developments on Monte Carlo Tree Search such as memory-augmented MCTS \cite{memorymcts} outperformed the original MCTS in the game of Go. Other methods like MCTSnets\cite{MCTSnets} incorporate simulation-based search inside a neural network, but it remain computationally very expensive.

\section{Methods}
\label{headings}
\subsection{Overview}
%
%
The Deep Pepper chess engine consists of a neural network (value network) which predicts the value of a given state as well as the optimal move (policy network) at a given state. The policy network is coupled with an MCTS algorithm, which performs a number of simulations of the most probable moves as dictated by an upper confidence tree selection process, initially based on the neural network's predictions. Coupling the neural network architecture with Monte Carlo Tree Search produces a powerful group of training algorithms called expert iteration\cite{Barbernp} in the framework of Classification-based Approximate Policy Iteration (CAPI)\cite{FarahmandPBG14}.  Unlike Alpha Zero\cite{ChessZero}, Deep Pepper is trained by embedding prior knowledge of the game of chess. Deep Pepper leverages the Stockfish engine for early stopping of games to accelerate training, and can be optionally pretrained using games from professional players and Stockfish evaluations. It also uses custom feature representation.

\subsection{Monte Carlo Tree Search}
Monte Carlo Tree Search (MCTS) is used by Deep Pepper in both policy evaluation and policy improvement. MCTS is a heuristic search algorithm used in Deep Pepper to  evaluate the most promising moves given the current board state. The algorithm is broken down into three main phases:
\begin{enumerate}
\item  \textbf{Selection:}  Starting from the root node of the current tree, the most promising child node, is selected until an unvisited node is encountered.
\item  \textbf{Expansion and Evaluation:} Once a leaf node is encountered, the leaf is expanded using Deep Pepper's neural network, which outputs move probabilities for all possible moves on the board, these probabilities are reduced to just legal moves. The value $v$ of the leaf node is computed from our value network.
\item \textbf{Back-up:} The statistics for each of the move in the explored tree are updated, including updating each of the action values with the output of the neural network, as well as the number of times each of the move has been taken.
\end{enumerate}

%
%
Deep Pepper stores each of the board states as a node, with each node containing information about the number times each of its children were visited $N(s,a)$ (where s is the state, and a the action), the total value (as determined by the neural network or game conditions) of all descendant leaf nodes expanded beneath each node $W(s,a)$, the average value for each node $Q(s,a) = \frac{W(s,a)}{N(s,a)}$ and the initial move probabilities, $P(s,a)$ (as determined by the neural network).  Additionally, Dirichlet noise is added to the initial probabilities of the root node during each call of MCTS, to further favor exploration and avoid aggressively repeatable play.

MCTS in Deep Pepper explores states and actions using a variant of Polynomial Upper Confidence Tree's (PUCT)\cite{rosin2011}. With PUCT, the selection process is based on estimates on the upper bound of selection probabilities $U(s,a)$ where $$U(s,a) = c_{puct} P(s,a) \frac{\sqrt{\sum_k N(s,k)}}{1+N(s,a)}$$

$c_{puct} $ is a hyper-parameter, called the exploration factor, controlling how much one would like to explore less visited nodes. In this variant, we also add the average value for each given node to the PUCT values. That is, given the current statistics stored within a node, the child node is selected via:
$$ A = argmax_{a} (Q(s,a) + U(s,a)) $$

At the end of each iteration (800 simulations of MCTS), MCTS provides a policy to be used in the current game state.  This policy is determined by: $$\pi (s,a) = \frac{N(s,a)}{\sum_a N(s,a)} $$ 
Optionally, a temperature parameter $\tau$ has been added to account for the exploration-exploitation paradigm. This has been annealed to an infinitesimal value after the opening phase of the game. The policy is now determined via: $$\pi (s,a) = \frac{N(s,a)^{\frac{1}{\tau}}}{\sum_a N(s,a)^ {\frac{1}{\tau}}} $$

\subsection{State Representation}

The value network (which is used in evaluating leaf nodes) and policy network (which is used in determining prior probabilities of every state node in MCTS) require a board representation to be given as input. Some recent methods like Alpha Zero use the complete board representation of dimensions 8x8x73. This primarily poses 2 problems: 1) It is harder to train as it takes lot of time and too many layers, owing to its giant input representation. 2) In chess, two similar-looking boards might have a completely different evaluation. Traditional convolutional neural networks (CNNs) are not equipped to handle this issue. (This can be understood by using the following intuition: Consider a CNN trained to classify the images as cats or dogs. If you modify one pixel in a cat image, it is still supposed to output it as 'cat' (unless it is an adversarial attack\cite{su2017one}). However, in chess, if the white queen is at 'f3' instead of 'g3' the evaluation would change from -500 to +500 (on centi-pawn scale\cite{centipawn}) and CNNs cannot be trained to predict such drastic changes in output with only minor changes in the input.

To counter the above problems, taking inspiration from Giraffe\cite{Giraffe}, we design a custom 353-dimensional feature representation for given board that is faster to train and immune to adversarial attacks (to be proven with further experiments and human evaluation). Additionally, we demonstrate that it is a good representation as we observe favorable results during pre-training. Deep Pepper uses embedded knowledge to accelerate the training of the chess engine over a "tabula rasa" system such as Alpha Zero.  The state representation used by Deep Pepper hence uses knowledge derived by human experts about the value of pieces, and whether each piece and square is currently threatened and defended, and many other features (as detailed in the appendix).

\subsection{Neural Network Architecture}

We experiment with multiple architectures. 

\begin{wrapfigure}{r}{0.5\textwidth}
  \begin{center}
    \includegraphics[width=0.42\textwidth]{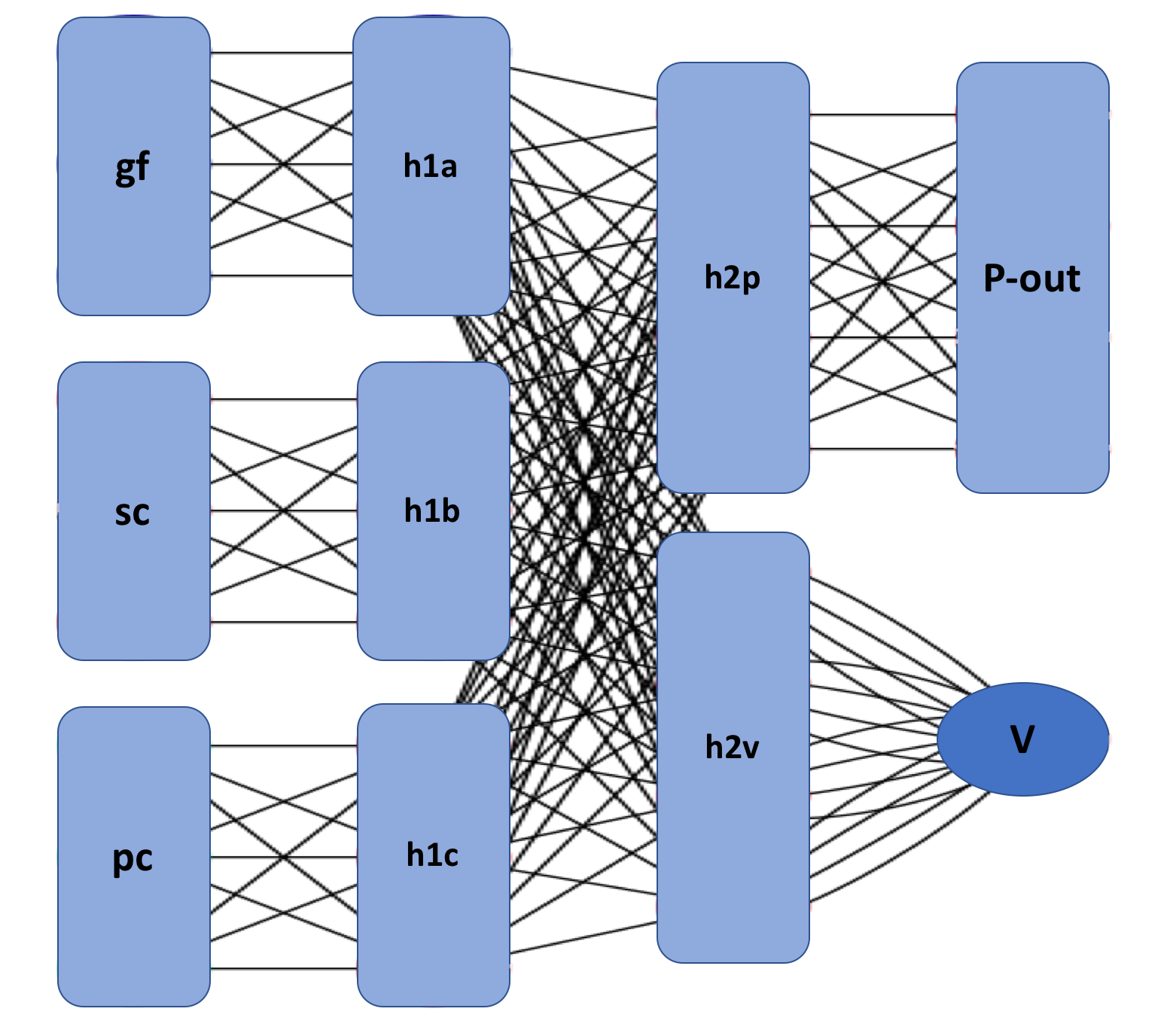}
  \end{center}
  \caption{Architecture of PolicyValNetwork-parts}
\end{wrapfigure}

\subsubsection{\textbf{PolicyValNetwork-parts}} 

From our feature representation, we connect our global features to h1a neurons(see appendix 7.12), piece centric features to h1b neurons and square centric features to h1c neurons in the first layer. We then concatenate all the outputs from first layer and connect to h2p neurons in the second layer and further this is connected to d-out neurons in the last layer which will give us the softmax policy output. As another head, we also connect the concatenated output from first layer to h2e neurons in the second layer which is further connected to 1 output neuron which gives us the value function. Therefore, we are using value and policy networks as different heads (with last 2 layers being different) in the same large network. This is much faster than the architecture with 2 separate networks, and without significant loss in performance. The number of different neurons used in each layer are detailed in appendix. 

 
 

\subsubsection{\textbf{PolicyValNetwork-Full}}

Similar to PolicyValNet-parts, but rather than having separate connections in the first layer, we have a fully connected layer which allows the network to learn more implicit connections between each of the global, square centric and piece centric features. An insignificant performance improvement is noticed which suggests that these extra connections are not effective considering the extra training time.

\subsubsection{PolicyNetwork-parts and Critic-parts}

Here, we use 2 separate networks for policy and value. The architectures are exactly similar to 3.4.1 (PolicyValNetwork-parts) except that, here the first layer is also trained independently for Policy and Value (critic) networks. This architecture has given a minor improvement in the performance.

\subsubsection{PolicyNetwork-full and Critic-full}

It is exactly similar to 3.4.2 except that, here the first layer is also trained independently for policy and value networks. As predicted, there is only a very slight improvement in the performance.

\subsection{Training Algorithm Pipeline}

Deep Pepper was trained using a self-play mechanism. In this format, the system learns a game primarily by playing against itself.  The training pipeline can be decomposed into thee main steps:
\begin{enumerate}
\item \textbf{Game generation}, wherein data is generated to be used to update the network
\item  \textbf{Network training}, wherein the network is improved using back propagation on the aforementioned generated games
\item  \textbf{Improvement testing}, wherein the new network is tested against the previous iteration in a series of games
\end{enumerate}

Monte Carlo Tree Search is used both during the game generation phase as well as the improvement testing phase to select the best move at each position.

Generally, one can view this training procedure as a form of classification-based approximate policy iteration (CAPI). The policy improvement step begins with MCTS performing a search based on the current network, $f_\theta$, which improves upon the current search policy. This new search policy is projected back into the network's function space via back-propagation during network training. Next, the policy evaluation step can be viewed as values of the outcomes of generated games, which are also projected back into the network's function space via back-propagation during each step network training. (similar reasoning is provided in the AlphaZero paper as well).

\subsubsection{Game Generation}
During the game generation phase of the training pipeline, Deep Pepper plays many games against itself, generating data to be used in the network training phase of the algorithm. During this phase, the current iteration of the network, $f_\theta$ is used in conjunction with MCTS to play games from start to finish. During each turn of the game MCTS is called, using the current state as the root node, determining the optimal policy for the current board state. The most probable move is selected as the best action. Furthermore, each of the states and policies, output by MCTS, are stored as expert policies (to be used for training the network). Next, MCTS is called using the next board state as the root node, and the current collected statistics from MCTS are stored and used again during the following batch of simulations. 

In order to accelerate training, Stockfish evaluations are used after $k$ half moves, if the Stockfish evaluation indicates a high probability for the current player to win/lose, the game ends with resignation by that player/opponent. Otherwise, the game continues until ending via draw, or checkmate. Once the game is over, the final value is added to each of the saved states and policies. The score at the end of the game is 0 in the case of a draw, +1 if white wins, or -1 if black wins. In the case of resignation, the back up values are positive if black resigned, or negative if white resigned, and are based on the Stockfish evaluation (with appropriate scaling) at the time of resignation.

\subsubsection{Network Training}

During this phase of the training pipeline, the updated policies and values are projected back into the function space $f_\theta$ defined by the neural network. The data created in the game generation phase is a set of triplets saved from each state of each game played. Each triplet, $\{s , \pi (s,a) , \nu \} $, where $s \in S$ is the state of the board, $\pi (s,a): S \rightarrow A $ is the policy generated from MCTS taken at the current board state,  and $\nu \in \mathbb{R}$ is the backed up end game value as determined via, draw, winner or resignation.  Each of these triplets is passed into the neural network,  with $\pi(s,a)$ acting as the label for the networks policy output and $\nu$ acting as the label for the value output of the network.

The objective function defined for the network is as follows:
$$J = L_{MSE} (f_\theta , z) + L_{CE} (f_\theta , \pi (s,a)) + \lambda \Omega(\theta ) $$
$$J = (f_{\theta val}(s) - z)^2 - \pi (s,a)^T log(f_{\theta pol}(s)) + \lambda || \theta ||^2$$
Where $L_{MSE}$ is the mean squared error loss, between the predicted and realized value of the state at the end of the game, $z$ . $L_{CE}$ is the cross entropy loss, for the learned policy, $\pi (s,a) $ and the predicted policy $f_\theta(s)$. And $\Omega(\theta)$ is the $L2$ regularization loss, applied to the network weights. Using KL-Divergence loss instead of $L_{CE}$ reminds us of distillation\cite{distillation}. The network is trained via back propagation with gradient updates calculated using an Adam optimizer on mini-batches.

\subsubsection{Improvement Testing}

Once the network parameters have been updated, there is an optional check to ensure that the new network has improved over the last iteration. In order to measure network improvements, self-play is again used as in game generation, however, the previous generation is paired against the new candidate. As before, at each turn of the game, MCTS guided by either the updated network or an older iteration, and used to determine the optimal move at every given turn. The game is played until checkmate, draw or resignation due to Stockfish evaluation.  We only save the outcome of each game for ranking, but no other data during this phase of training.

Should the updated network defeat the older iteration by more than half the time, the newly trained network is kept and used in the next round of game generation.

\subsection{Pretraining}

An optional pretraining phase was added to some models in order to offer early guidance in training the network.  During pretraining, board positions (taken from professional games of chess), are assigned value by the Stockfish engine. Next, an approximate policy label is generated using the Stockfish engine to compute a value for each possible subsequent board, and calculating an approximate policy by applying the softmax function to the set of generated values. This is because, as detailed before, training of the policy network directly on expert games does not work.

$$\pi_{approx}(s,a)= (\frac{\exp{(Stockfish(s_1))}}{ \sum_k \exp{Stockfish(s_k) } }, \cdots,\frac{\exp{(Stockfish(s_j))}}{ \sum_k \exp{Stockfish(s_k) } }) $$

The generated labels are used to train the networks via supervised learning. In this way, the networks are capable of initially analyzing an approximate value for some common board positions. The same cost function is used as during the main training pipeline:
$$J = L_{MSE} (f_\theta , \nu) + L_{CE} (f_\theta , \pi (s,a)) + \lambda \Omega(\theta ) $$
which is optimized again using an Adam optimizer.

Overall, pretraining was performed on approximately 100,000 board positions seen in games played by Kasparov, Karpov, Ivanchuk, Aronian and others.

\section{Results}

\begin{wrapfigure}{r}{0.5\textwidth}
 \label{figure-2}
  \begin{center}
    \includegraphics[width=0.42\textwidth]{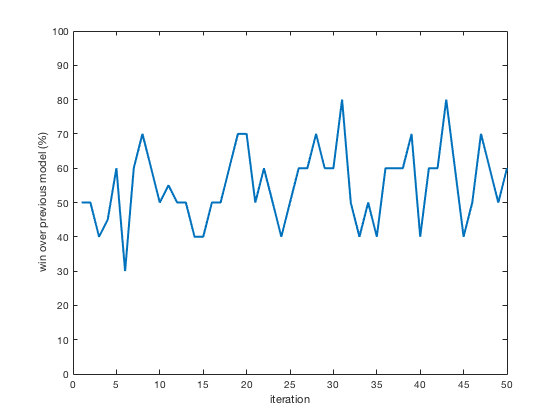}
  \end{center}
  \caption{Performance of the current model against it's previous iteration, tested on 10 games at every iteration as the training progresses. We can see that, about 70 percent of the time, our current model is observed to better than it's previous iteration.}
\end{wrapfigure}

Results can be seen in Table \ref{results-table}. Experiments were run by testing trained networks against a purely random decision playing process. So far, only three models were tested in this fashion. Each model was pretrained on a set of approximately one hundred board positions, then trained for some number of games afterwards. Pretrain0 refers to a model that was just pretrained, Pretrain50 refers to a model that received 50 games of additional training and so on.

Additionally, models were tested against previous iterations from their own evaluation. Models were shown to improve as training continued as shown in \ref{figure-2}. A model that was trained for 100 iterations (and had pretraining) beat a purely pretrained model in 56 percent of games. Further, a pretrained model won against a randomly initialized model in 64\% of games.

\begin{table}[h]
  \caption{Trained networks versus random play on 25 games}
  \label{results-table}
  \centering
  \begin{tabular}{lll}
    \toprule
    \multicolumn{2}{c}{Part} \\
    \cmidrule(r){1-2}
   Model     &  Win-rate    \\
    \midrule
    Pretrain0 &    \\
    Pretrain50     & 0.55       \\
    Pretrain100     &    0.64      \\
    Pretrain150 & 0.64 \\
    \bottomrule
  \end{tabular}
\end{table}

\section{Future Work}

\subsection{Improved Evaluation Mechanisms}

When developing a complex system of this nature, evaluation is a critical step to ensuring the system is behaving as expected and meeting desired performance. At this time, Deep Pepper has been tested only against past models, and models that have had extensive pretraining. Naturally, this limits the scope of evaluation to determining whether the system is improving, but prevents analysis of objective metrics such as ELO. For this reason, a priority of future work will focus on expanding evaluations of models to include ELO rating as an objective metric of skill. 

Should ELO based metrics be challenging to test effectively it would also be of interest to test current iterations against past iterations using win percentage as a gauge for growth in the systems skill. This system could be quickly implemented to run parallel to the current system and continuously play a subset of networks against each other, as the network is updated, iterations can be added to the subset of models for evaluation, to show estimated changes in win rates based on number of iterations.

\subsection{System Optimization}
The training pipeline in its current form could be accelerated in a number of ways to reduce the time taken at the primary bottleneck, game generation. More concretely, MCTS takes significant time due to it's high usage of computation resources for determining all legal moves for each board, and calculating $f_{\theta pol}$ and $f_{\theta val}$. These processes could be further optimized by having a large number of processes  running simultaneously during game generation. Once a process reaches the expansion phase of the MCTS process, it would stop and wait for a number of other processes to reach the same point. This would reduce the computation cost in using the neural network (as the underlying framework, Pytorch, can be optimized to use a GPU to compute the outputs on a large number of samples).

Training time could also be saved by having each separate MCTS process share node statistics among each other. This change would reduce the time spent expanding identical board positions, which would now be shared among processes, but the impact on total convergence rate of MCTS is yet to be explored.

\subsection{Hyper-parameter Selection and Network Optimization}
At this time, very little hyper-parameter optimization has been performed on the system, leaving room for improvement, since a system of this nature has many hyper-parameters. Initial values for hyper-parameters such as number of MCTS simulations, and PUCT exploration factor were taken from Alpha Zero \cite{ChessZero} and Alpha Go \cite{AlphaGo}. There is no evidence that these values are still optimal given the variation between Deep Pepper and these systems. Additionally, other hyper-parameters, such the number of games to generate between network training have been set to values that have been observed to work empirically, but extensive search for an upper limit of the number of games to use in each epoch has not been performed.  Extensive testing should be done to determine the available range of games which can be generated, and the optimal value for maximizing the rate at which the network learns to play.

Furthermore, while several network architectures have been tested, there has not been extensive testing on all available architecture options. Batch normalization\cite{batchnorm} has the potential to rapidly increase the speed at which the network learns to play. It would also be of interest to explore alternative activation functions like swish \cite{Swish}, as well as better optimization techniques like \cite{newAdam}.



\section{Acknowledgements}


The work was done as a course project for Prof. Doina Precup and Pierre-Luc's Reinforcement Learning course taught at McGill University. The authors would like to thank Doina Precup, Pierre-Luc, Rithesh Kumar, Sherjil Ozair, Kyle Kastner,  Vidhi Jain, Giancarlo Kerg, Zihan Wang for useful feedback and discussions. This work utilized the computing facilities  managed  by  the  Montreal  Institute  for  Learning  Algorithms,  NSERC, Compute Canada, and Calcul Quebec. The authors would also like to express a debt of gratitude towards those who contributed to the open source projects Stockfish and Giraffe.

\bibliographystyle{plain}
\bibliography{main}

\section{Appendix}

\subsection{Features and Parameters}

\subsubsection{Feature vector}
%
This is a slightly modified version of our interpretation of Giraffe's\cite{Giraffe} feature representation.

\textbf{0:} Side to move: 1 for White and 0 for Black 

\textbf{1-4:} Binary values for castling rights (White queen-side castling, White king-side castling, Black queen-side castling, Black king-side castling in that order).

\textbf{5-16:} Number of pieces of each type. White King, White Queen(s), White rooks, White Bishops, White knights, white pawns, black king, black queen(s), black rooks, black bishops, black knights, black pawns in that order.

\textbf{17-176:} For every piece, we determine whether the piece is existent (1 slot), normalized x-coordinate (1 slot), normalized y-coordinate (1 slot), lowest valued attacker of that piece (1 slot), and lowest valued defender of that piece (1 slot). So, we use 5 slots for every piece. And, there are a total of 32 pieces. (2 kings, 2 queens, 4 rooks, 4 bishops, 4 knights, 16 pawns). So, a total of 160 slots.

\textbf{177-224:} For each major and minor piece, this encodes how far they can go in each direction. So 8 slots for a queen, 4 slots for a rook and 4 slots for a bishop. Since there are 2 queens, 4 rooks and 4 bishops, we need a total of 48 slots.

\textbf{225-352:} For each square, we encode the values of the lowest valued attacker and lowest valued defender of that square. Since we have 64 squares, there are a total of 128 slots used here. 

\subsubsection{Network parameters}

d-in = 353 (size of input feature representation)

global-features = 17 (includes slots 0-16 described in section 9.1)

piece-centric = 218 (includes slots 17-224 described in section 9.1)

square-centric = 128 (includes slots 225-353 described in section 9.1)

h1a = 32 (no.of neurons in the first hidden layer to be connected to the global features in the input layer for Policy-parts and critic-parts networks)

h1b = 512 (no.of neurons in the first hidden layer to be connected to the piece-centric features in the input layer for Policy-parts and critic-parts networks)

h1c = 480 (no.of neurons in the first hidden layer to be connected to the square-centric features in the input layer for Policy-parts and critic-parts networks)

h2p = 2048 (no.of neurons in the second hidden layer which are fully connected to all the neurons in the first hidden layer in Policy head of PolicyValNetwork-parts and PolicyNetwork-parts)

h2e = 512 (no.of neurons in the 2nd hidden layer which are fully connected to all the neurons in the first hidden layer in Value head of PolicyValNetwork-parts and Critic-parts networks)

h1 = 1024 (neurons in the first hidden layer for policy-full and critic-full networks)

h2 = 2048 (neurons in the second hidden layer for policy-full and critic-full networks)

d-out = 5120 for policy outputs. It is represented as "from square" to "to square" which gives us 64x64 = 4096 possibilities and the additional neurons are added to account for pawn promotions and other miscellaneous moves.


\subsection{Mathematical Details}

\subsubsection{UCT}

These are inspired from Upper Confidence Bound (UCB) based algorithms for k-arm bandit problems \cite{ucb-bounds} . The Upper Confidence Tree (UCT) based algorithm was first introduced in \cite{uct}. The objective is to find a path from the root node to the leaf node which maximizes the mean reward defined as $Q_i = E[r_i]$. For non-leaf nodes, the Q-value is defined as $Q_k = max_{i \in T(k)} Q_i$, where, $T(k)$ is the set of all leaf nodes in the subtree with root $k$.

In each iteration $t$, a node is selected (until a leaf node is encountered) which maximizes the $A$-value defined as:
$$ A_t(k) = \hat{Q}_{k,t} + c   \sqrt[]{\frac{log N_j(t)}{N_k(t)}} $$
where, c is a positive constant, $N_k(t)$ is the number of times the node $k$ was visited, $N_j(t)$ is the number of times the parent of node $k$ was visited. Note that, in the case of MCTS, each child node is the corresponding state-action pair, and hence $ \hat{Q}_{k,t} $ is exactly $Q(s,a)$ where we ignore the iteration number $t$ for convenience in representation.
So, 
$$ \hat{Q}_{t,k} = \frac{1}{ N_k(t)} \sum_{s=1}^t r_s \mathds{1} [I_s \in T(k)] $$ 
This can be compared with:
$$ Q(s,a) = \frac{W(s,a)}{N(s,a)} $$

Theoretical guarantees for the lower bound on the regret of UCT are given in \cite{uct}. But, the worst case analysis given in \cite{coq} shows that it could be worse than uniform sampling. 

\subsubsection{Classification Based Approximate Policy Iteration}

The formulation of deep pepper fits into the broader class of algorithms called "Classification based Approximate Policy Iteration" (CAPI) introduced in \cite{FarahmandPBG14} which is given below:

\hrulefill

\textbf{Algorithm} CAPI($\Pi$, $\nu$, $K$)

\indent\textbf{Input:} policy space $\Pi$, state distribution $\nu$, number of iterations $K$\\
\indent\textbf{Initialize:} Let $\pi_{(0)} \in \Pi$ be an arbitrary policy\\
\indent\textbf{Loop:} for $\mathbf{k} = 0,1,\ldots ,K$ \textbf{do}
\indent\begin{itemize}
\item construct a dataset $\mathcal{D}_n^{(k)} = \{X_i\}_{i=1}^n$, with $X_i \overset{\mathrm{i.i.d.}}{\sim} \nu$
\item $\widehat{Q}^{\pi_k} \leftarrow \textrm{PolicyEval}(\pi_k)$
\item $\pi_{k+1} \leftarrow \textrm{argmin}_{\pi \in \Pi} \widehat{L}_n^{\pi_k}(\pi)$ (approximate policy improvement step with action-weighted classification)
\end{itemize}
\hrulefill

where, the action-gap weighted empirical loss function $\widehat{L}_n^{\pi_k}(\pi)$ in policy space $\Pi$ is defined as:
\begin{align*}
\widehat{L}_n^{\pi_k}(\pi) &\overset{\mathrm{def}}{=} \int_{\mathcal{X}} \mathbf{g}_{\widehat{Q}^{\pi_k}}(x) \cdot \mathbb{1}\{\pi(x) \neq \textrm{argmax}_{a\in \mathcal{A}} \widehat{Q}^{\pi_k}(x,a) \} d\nu_n \\
                                         &= \sum_{X_i \in \mathcal{D}_n^{(k)}} \mathbf{g}_{\widehat{Q}^{\pi_k}}(X_i) \cdot \mathbb{1}\{\pi(X_i) \neq \textrm{argmax}_{a \in \mathcal{A}} {\widehat{Q}^{\pi_k}}(X_i,a) \} 
\end{align*}

and, the action-gap for a MDP with 2 actions is defined as: 

$$\mathbf{g}_{Q}(x )\overset{\mathrm{def}}{=} |Q(x,1)- Q(x,2)|$$ for all $x \in \mathcal{X}$

By using loss distortion lemma, they provide theoretical guarantees for the convergence of CAPI algorithm. They further give directions on extending this to MDPs with multiple actions in each state. By seeing Deep Pepper as an approximation of CAPI (see section 3.5), similar guarantees for convergence can be proven.  



\end{document}